# Efficient DWT-based fusion techniques using genetic algorithm for optimal parameter estimation

**Kavitha S**
Department of Computer Science and Engineering
SSN College of Engineering, Chennai – 603110, India.
E-mail: kavithas@ssn.edu.in

**Thyagharajan K K**
Department of Electronics and Communication engineering
RMD Engineering College, Chennai – 601206, India
E-mail: kkthyagharajan@yahoo.com

**Abstract** Image fusion plays a vital role in medical imaging. Image fusion aims to integrate complementary as well as redundant information from multiple modalities into a single fused image without distortion or loss of information. In this research work, discrete wavelet transform (DWT)and undecimated discrete wavelet transform (UDWT)-based fusion techniques using genetic algorithm (GA)foroptimalparameter(weight)estimationinthefusionprocessareimplemented and analyzed with multi-modality brain images. The lack of shift variance while performing image fusion using DWT is addressed using UDWT. The proposed fusion model uses an efficient, modified GA in DWT and UDWT for optimal parameter estimation, to improve the image quality and contrast. The complexity of the basic GA (pixel level) has been reduced in the modified GA (feature level), by limiting the search space. It is observed from our experiments that fusion using DWT and UDWT techniques with GA for optimal parameter estimation resulted in a better fused image in the aspects of retaining the information and contrast without error, both in human perception as well as evaluation using objective metrics. The contributions of this research work are (1) reduced time and space complexity in estimating the weight values using GA for fusion (2) system is scalable for input image of any size with similar time complexity, owing to feature level GA implementation and (3) identification of source image that contributes more to the fused image, from the weight values estimated.

**Keywords** Image fusion · Genetic algorithm · Discrete wavelet transform· Undecimated discrete wavelet transform, Objective metric.

## 1 Introduction

Advancements in technology have revolutionized almost every aspect of medical imaging. With the rapid developments in high technology and modern instrumentation, medical image fusion has become a vital aid for medical diagnosis, treatment and research. Medical imaging is the process, which produces images of internal aspects of the body by either invasive or non-invasive techniques. To support more accurate clinical information, medical images are required by the physicians for diagnosis and treatment (Goshtas and Nikolov 2007; Dammavalam et al. 2012).

In the field of medical image processing and analysis, radiologists require high-resolution medical images with information such as region, tissue and visualization to help with improved disease diagnosis and computer assisted surgery (National Brain Tumor Society 2015; American Brain Tumor Association 2015). These requirements cannot be resolved with single modality medical images, because each of the imaging technique is designed to capture only specific aspects of the human anatomy. Computed tomography (CT) is more popularly used for recognizing the bone structure and tumor region, the soft tissue information is more visible in magnetic resonance image (MRI), positron emission tomography (PET) is useful in the diagnosis of brain disease, brain tumors, strokes, and neuron-damaging diseases (dementia) while single photon emission computed tomography (SPECT) conveys clear information in blood flow analysis during active/inactive state of the brain (American Brain Tumor Association 2015). For efficient disease diagnosis, complementary information from multiple modalities becomes necessary (Nishele 2015). Thus, fusion of multimodality medical images has become a promising and very challenging research area in recent years (Wang et al. 2005; Brainimages&information2015).This research work focuses on designing a fusion system for complementary information retrieval and analysis for the images acquired from multiple sensors of the patient during nearly same timeframes.

**1.1 Image fusion and its classification**

Image fusion is defined as a process of combining multiple input images or some of their features into a single image without introducing distortion or loss of information (Krishnamoorthy and Soman 2010; National Brain Tumor Society 2015). Image fusion techniques are broadly classified as pixel level, region level and feature level techniques (Rajkumar and Kavitha 2010). In the past, mostly pixel level fusion techniques were used (Fusion tool 2015; Shah et al. 2013). In the next variant, the image is decomposed into regions and then pixel level fusion is applied. Feature level techniques are widely used in data fusion process rather than image fusion process (Khaleghi et al. 2013).

Fusion techniques are also classified based on spatial (pixel) and transformation methods. The recent spatial fusion methods are related to machine learning techniques such as neural networks and fuzzy approach. The transformation fusion methods widely used are pyramid, contour and wavelet-based techniques. To optimize the fused image for better contrast and information, genetic algorithm (GA) is used (Majid et al. 2008).

*1.1.1 Spatial fusion techniques*

Neural network is adapted as pulse coupled neural network (PCNN) algorithm (Wang and Ma 2008; Wang et al. 2010) and its variants like multi-channel PCNN, dual-channel PCNN are used in fusion for images from different domains such as satellite, infrared and medical images that have been taken from multiple sensors. Fuzzy logic, which has been used for image fusion over the past two decades, still plays a vital role in fusion research, since it is practiced for processing uncertain and ambiguous data derived in real-time applications. For image uncertainty, the membership functions are used, to describe the distribution and clustering of the pixel values, to derive fusion operators and decision rules for image fusion. Therefore, image fusion rules (decision rules) based on fuzzy logic uses fuzzy inference procedure to solve the issue of uncertainty in the images (Teng et al. 2010; Singh et al. 2004; Teng et al. 2010).

*1.1.2 Transformation fusion techniques*

The pyramid-based methods of fusion are filter-subtraction decimate pyramid (FSD), gradient pyramid, Laplacian pyramid, morphological pyramid, ratio pyramid, contrast pyramid, etc. Each of these approaches has its own limitation in the fusion process. For example, contrast pyramid method loses too much information from the source images; ratio pyramid method produces false information that does not exist in the source images and morphological pyramid method creates multiple false edges (Krishnamoorthy and Soman 2010; Fusion tool 2015). These methods are hence not ideally suited for the fusion of medical images.

Another widely used transformation method is region based contourlet transform which retains localization, directionality, anisotropy, etc., in the fused image. It is generally implemented in two stages, transformation and decomposition but here the computational complexity is high (Zhang andBao-Long2009;Yangetal.2008;RajkumarandKavitha 2010). To sum up, we could say that, all the methods discussed above have some limitation or the other—in preserving the texture information, background, contrast and edge from the source image during fusion.

Hence, the majority of fusion algorithms are based on wavelet transformation (Pajares and de la Cruz 2004; Zheng et al. 2007). In discrete wavelet transform (DWT), downsampling is done during decomposition and upsampling is done in reconstruction step. However, the DWT in medical image fusion results with shift variance. To overcome the problem of shift variance, undecimated discrete wavelet transform (UDWT) is used. In UDWT, only upsampling operation is used (Raj and Venkateswarlu 2011).

**1.2 Genetic algorithm (GA) in fusion**

Genetic algorithm is used in various optimization problems such as function optimization, parameter optimization of controllers, multi objective optimization, fused image optimization etc. (Homaifaretal.2010;JagadeesanandParvathi 2014; Kaur et al. 2014). The basic steps of GA are initial population design, selection, crossover, mutation and fitness function evaluation to select the optimal subsets for the next generation. In this paper, to construct the optimal fused image, weight factor for each of the source images is computed and applied during fusion. In Majid et al. (2008); Dammavalam et al. (2012), a pixel level weighted average image fusion based on GA has been done for infrared and visible source images and validated with quantitative metrics like Image Quality Index (QI), mutual information (MI) and root mean square error (RMSE). The drawback of this method is its computational cost of performing fusion, since each pixel of source image is considered during initial population to find the computed weight values. To overcome this drawback of computational cost, we propose a feature-based weight computation genetic function using mean squared error (MSE) as the fitness function for evaluation and analysis.

From recent literature available on this topic, the importance and progress of optimal parameter estimation in real-time scenario has been studied. In Jiang and Wu (2015), wavelet transform (WT), hidden Markov model (HMM) and genetic algorithm (GA) are used in sequence for the fusion of multi-scale images with weight factor and evaluation shows that the information, texture and edge of the fused image are improved. The algorithm using HMM and GA for fusion is illustrated; however, the computed weight value is not mentioned in the results. In addition, the weight factor is calculated and used in various applications like exposure fusion weight for the images of under or over exposed pixels of grey and color type in fusion process (Moumene et al. 2014), optimal feature selection and compression of feature space using GA for content-based image retrieval (Chandrashekhar et al. 2015), fusion with GA is used in combining the features of thermal and visible information efficiently for the selection of optimal face areas which in turn improves the face recognition accuracy (Hermosilla et al. 2015).

In this paper, DWT-based fusion techniques using GA for optimal parameter estimation are proposed for the fusion of multimodality brain images. Further the fused image has been validated using subjective and objective metrics. Subjective metric is a human perception and the objective metric is the statistical evaluation. The objective metrics chosen for our work from (Naidu and Raol 2008) are: the impact of information using information entropy (IE), to measure the information deviation from source to the fused image using MI and QI, noise of the fused image with peak signal to noise ratio (PSNR), error between the source to the fused image with RMSE and the overall spatial activity information using spatial frequency (SF). The main contributions of this work are (1) minimization of search space, (2) scalability of the algorithm to handle input image of any size (3) identification of the source image that contributes more medical information to the fused image.

The paper spans the following sections: Sect. 2 elaborates the design of the proposed system; Sect. 3 describes the chosen evaluation criteria based on the quantitative metrics; Sect. 4 illustrates the experimental results and inferences of the proposed model on fusion techniques. Conclusion and Discussion are stated at the end.

## 2 System design

In this paper, multimodality fusion techniques are proposed using DWT and UDWT. These techniques are modified with GA for optimal weight estimation to obtain a better fused image. Then the fusion results from DWT, UDWT, DWT-GA, UDWT-GA are evaluated and analyzed using the objective metrics such as IE, MI, QI, RMSE, PSNR and SF. The overall system design is shown in Fig. 1.

### 2.1 Dataset

The brain images used in the analysis are acquired through different modalities such as SPECT (SPECT-Ti and SPECTTc), MRI (MRI-T2 and MRI-GAD) and CT and have been collected from Harvard medical school website (Johnsonand Becker 1999). The combination of two different modality images of each dataset, taken for the same patient during the same time is shown in Table 1. The images with different sizes are registered to the same size of one another, before the fusion process.

## 2.2 Discrete wavelet transform (DWT)

The DWT algorithm is implemented in two stages. The first stage involves image decomposition and the second stage involves fusion using inverse DWT (Raj and Venkateswarlu 2011).

*Stage I*

- Read the two source images A and B to be fused.

- Perform independent wavelet decomposition of two images until level two. Wavelet decomposition includes filtering (using low and high pass filters) and down sample by a factor of 2. Here db1 filter is used, since it retains the contrast and smoothness of the image. The decomposed matrices are LL-approximation, LH-vertical, HL horizontal and HH-diagonal.
- Apply the fusion decision rule (Maximum).

**Fig. 1** Proposed system design

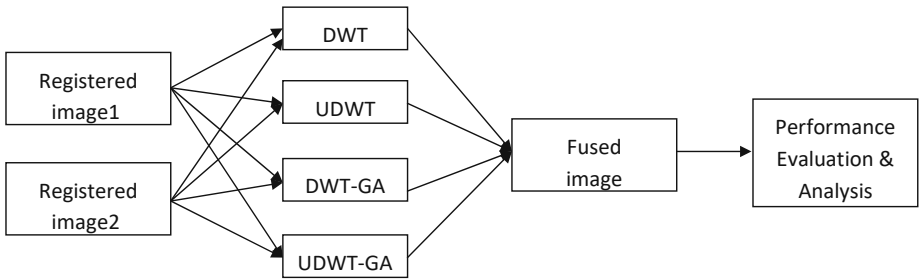

**Table 1** Dataset with size and entropy value

| Dataset | Source image1 | Source image2 | Size of image1 | Size of image2 | Entropy of image1 | Entropy of image2 |
|---|---|---|---|---|---|---|
| 1 | MRI-T2 | SPECT-Ti | 300 × 300 × 3 | 256 × 256 × 3 | 4.075709 | 2.729425 |
| 2 | MRI-T2 | SPECT-Tc | 256 × 256 × 3 | 256 × 256 × 3 | 4.029973 | 4.372145 |
| 3 | MRI-Gad | SPECT-Ti | 300 × 300 × 3 | 256 × 256 × 3 | 4.896195 | 1.168748 |
| 4 | CT | MRI-T2 | 256 × 256 × 3 | 256 × 256 × 3 | 4.290422 | 6.147892 |

From the decomposed image, feature vectors (median, standard deviation, variance and seven normalized central moments) are computed and are given as input to the genetic algorithm for weight estimation (FeaImage1 for image 1, FeaImage2 for image 2).

*Stage II*

- The final fused transform LL corresponding to approximations through pixel rules and the vertical, horizontal and diagonal details (LH, HL and HH) are obtained. Image synthesis includes upsampling by a factor 2 followed by filtering (low and high pass).
- The new coefficient matrix is obtained by concatenating the fused approximations and details.
- Fused image is reconstructed using inverse wavelet transform and displayed.

The result of this stage is the output of DWT_IF.

## 2.3 Undecimated discrete wavelet transform (UDWT)

Discrete wavelet transform is not shift invariant, hence it leads to artifacts in the fused images. To preserve shift variance property, undecimated discrete wavelet transform is used (Raj and Venkateswarlu 2011).

This algorithm is implemented in two stages. First stage involves image decomposition and the second stage involves fusion using inverse UDWT. *Stage I*

- Read the two source images A and B to be fused.
- Perform independent wavelet decomposition of the two images until level two. Wavelet decomposition includes filtering (using low and high pass filters). Here db1 filter is used, since it retains the contrast and smoothness of the image. The decomposed matrices are LL-approximation, LH-vertical, HL-horizontal and HH-diagonal.
- Apply the fusion decision rule (Maximum).

From the decomposed image, feature vectors (median, standard deviation, variance and seven normalized central moments) are computed and are given as input to the genetic algorithm for weight estimation (FeaImage1 for image 1, FeaImage2 for image 2). *Stage II*

- The final fused transform LL corresponding to approximations through pixel rules and the vertical, horizontal and diagonal details (LH,HL and HH) are obtained. Image synthesis includes upsampling by a factor 2 power $(j-1)$ where $j$ represents level $j$ followed by filtering (low and high pass).
- The new coefficient matrix is obtained by concatenating the fused approximations and details.
- Fused image is reconstructed using inverse wavelet transform and displayed.

The result of this stage is the output of UDWT_IF.

## 2.4 Modified GA for optimal parameter (weight) estimation

The features extracted from the two source images of DWT/UDWT decomposition are given as input to the modified GA, from which an optimal weight value for the fusion process is estimated and applied to the DWT/UDWT-based fusiontechniques.ThealgorithmofmodifiedGAforoptimal weight estimation is given below:

### 2.4.1 Algorithm

1. The feature vectors of size (1*40), computed from theregistered source images (after decomposition) are consideredasinitialpopulation—*FeaImage*1, *FeaImage*2
2. Initialize the variables : w*v* = 0, w*t* = 1 (such that w*v* + w*t* = 1), *diff* = 0.1, *generation* = 100, *MSE* = 0, *n* = 10 where

    w*v*, w*t* are optimal weight values *diff* is used as a mutation parameter to re-compute the weight value for the next generation *generation* is the number of generations
    *MSE* is the fitness value of each individual initially assigned to zero *n* is the number of trial runs in each iteration

3. To compute the fitness value using *MSE*, three parameters are considered: *FeaImage*3, *FeaImage*1 and *FeaImage*2 (each with size 1*40). Here *FeaImage*1 and *FeaImage*2 are features of two input images after decomposition and *FeaImage*3 is computed as given in Eqs. 1 to 4.

    $$FeaImage3 = D.*m1 + (\sim D).*m2 \qquad (1)$$
    $$m1 = wv * FeaImage1 \qquad (2)$$
    $$m2 = wt * FeaImage2 \qquad (3)$$
    $$D = (abs(m1) - abs(m2)) \qquad (4)$$

    Now the *MSE* is computed using the Eq. 5.

    $$MSE = \frac{1}{2}\left(\frac{1}{d}\sum_{i=1}^{d}[FeaImage3(i) - FeaImage1(i)]^2 + \frac{1}{d}\sum_{i=1}^{d}[FeaImage3(i) - FeaImage2(i)]^2\right) \qquad (5)$$

    Store the value of *MSE* in an array with current weight values.

4. Mutate the weight value using the *diff* variable as given in Eqs. 6 and 7.

    $$wv = wv + diff \qquad (6)$$
    $$wt = wt - diff \qquad (7)$$

5. Repeat steps 3 and 4 for *n* times and store the *MSE* value in an array.
6. Sort the *MSE* array and find the difference between two consecutive values. If it is less than 0.0001 terminate the process, or if the number of generations reaches the value of maximum generations, goto step 8. Once the termination condition is met, w*v* and w*t* at which the *MSE* value is minimum are chosen as optimal weights else goto step 7.
7. Re-compute the weight initialization and *diff* for the next generation as mentioned in Eqs. 8 to 10 and goto step 3.

    $$wv = wv((sortedmse\,[1]) - (diff/2)) \qquad (8)$$
    $$wt = wt((sortedmse\,[1]) + (diff/2)) \qquad (9)$$
    $$diff = diff/10 \qquad (10)$$

8. Return the optimized parameter (weight) value of each image as w*v*, w*t* respectively.

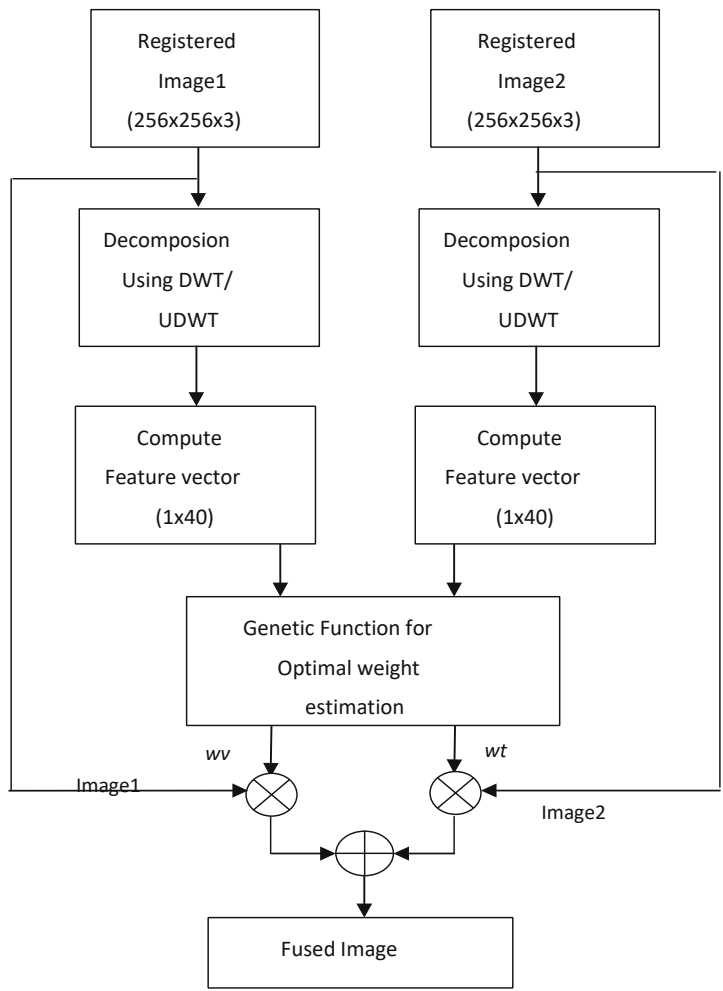

**Fig. 2** DWT/UDWT with modified GA for optimal weight estimation and fusion

**2.5 DWT/UDWT fusion technique with genetic algorithm (GA)**

The source images from two different modalities are registered and decomposed using DWT/UDWT. Then the feature vectors are computed from the decomposed images and given as input to modified GA, from which optimal weight values are generated (*wv*, *wt*) is explained in Sect. 2.4.1. The input images are updated with the obtained weight values and then the fusion rule (additive) is applied to generate the fused image. The entire process is shown in Fig. 2.

**3 Evaluation criteria**

The techniques DWT_IF, UDWT_IF, DWT-GA_IF and UDWT-GA_IF are compared using objective metrics such as IE, QI, MI, RMSE, PSNR and SF (Naidu and Raol 2008). The metrics chosen for analysis evaluates information, quality, error, noise and overall activity of the fused image.

Information entropy (IE): Information entropy is a statistical measure of randomness that can be used to characterize the texture of the input image. It is measured using Eq. 11. A higher value for the entropy signifies better quality of the fused image.

$$IE = -\sum_{i}^{L-1} p_i \log_2 p_i \qquad (11)$$

where $L$ is the number of grey levels and $p_i$ is calculated as given in Eq. 12.

$p_i$ = number of pixels $D_i$ of each graylevel $i$ / number of pixels $D$ in the image   (12)

Mutual information (MI): The MI measures the amount of information obtained from the fusion of input images. It is measured using equations mentioned in 13–15. Higher MI value indicates that, better fusion results are obtained.

$$MI = I_{FV}(F, V) + I_{FT}(F, T) \qquad (13)$$

where

$$I_{FV}(F, V) = \sum_{F,V} P_{F,V}(F, V) \log_2 \frac{P_{FV}(F, V)}{P_F(F) P_V(V)} \qquad (14)$$

$$I_{FT}(F, T) = \sum_{F,T} P_{F,T}(F, T) \log_2 \frac{P_{FT}(F, T)}{P_F(F) P_T(T)} \qquad (15)$$

where $V$ and $T$ are the input images. $F$ is the fused image. $P_V(V)$, $P_T(T)$ and $P_F(F)$ are the histograms of the images $V$, $T$ and $F$, respectively. $P_{FV}(F, V)$ and $P_{FT}(F, T)$ are the joint histograms.

Root mean square error (RMSE): The RMSE is computed from the differences between input image pixel values and the fused image pixel values, as mentioned in Eq. 16. Lesser value of RMSE indicates good quality of fused image.

$$RMSE = \frac{1}{2} \left( \sqrt{\frac{1}{MN} \sum_{i=1}^{M} \sum_{j=1}^{N} [F(i, j) - V(i, j)]^2} + \sqrt{\frac{1}{MN} \sum_{i=1}^{M} \sum_{j=1}^{N} [F(i, j) - T(i, j)]^2} \right) \qquad (16)$$

where $V(i, j)$ and $T(i, j)$ are the input images. $F(i, j)$ is the fused image. $M$ and $N$ are the number of rows and columns in the input images.

Peak signal to noise ratio (PSNR): PSNR is defined as the ratio between the maximum possible peak of a signal and the peak of corrupting noise that affects the fidelity of its representation. It is measured using Eq. 17. The signal in this case is the original data, and the noise is the error introduced by fusion.

$$\text{PSNR} = 10\log_{10} \frac{\max(I)^2}{MSE} \quad (17)$$

where $MSE$ is the mean square error and $I$ is the maximum possible pixel value.

Universal Quality Index (QI): QI measures the amount of salient information that has been transformed to the fused image from input images as shown in Eqs. 18 to 24. The range of this metric is $-1$ to $+1$

$$QI = \frac{\sigma_{xy}}{\sigma_x \sigma_y} * \frac{2\bar{x}\bar{y}}{(\bar{x})^2 + (\bar{y})^2} * \frac{2\sigma_x \sigma_y}{\sigma_x^2 + \sigma_y^2} \quad (18)$$

$$QI = \text{Correlation} \times \text{luminance} \times \text{contrast} \quad (19)$$

$$\bar{x} = \frac{1}{N} \sum_{i=1}^{N} x_i \quad (20)$$

$$\bar{y} = \frac{1}{N} \sum_{i=1}^{N} y_i \quad (21)$$

$$\sigma_x^2 = \frac{1}{N-1} \sum_{i=1}^{N} (x_i - \bar{x})^2 \quad (22)$$

$$\sigma_y^2 = \frac{1}{N-1} \sum_{i=1}^{N} (y_i - \bar{y})^2 \quad (23)$$

$$\sigma_{xy} = \frac{1}{N-1} \sum_{i=1}^{N} (x_i - \bar{x})(y_i - \bar{y}) \quad (24)$$

where $x$ is the input image1/image2 and $y$ is the fused image.

Spatial frequency (SF): Spatial frequency indicates the overall activity level (row wise, column wise) of the fused image in spatial domain and it is calculated as shown in Eqs. 25 to 27.

$$SF = \sqrt{RF^2 + CF^2} \quad (25)$$

$$RF = \sqrt{\frac{1}{MN} \sum_{X=0}^{M-1} \sum_{Y=1}^{N-1} [I_f(X, Y) - I_f(X, Y-1)]^2} \quad (26)$$

$$CF = \sqrt{\frac{1}{MN} \sum_{Y=0}^{N-1} \sum_{X=1}^{M-1} [I_f(X, Y) - I_f(X-1, Y)]^2} \quad (27)$$

## 4 Experimental results and analysis

The fused images obtained from the source images of different modalities taken for brain (as mentioned in Table 1), using four different types of fusion techniques, DWT_IF, UDWT_IF, DWT-GA_IF, UDWT-GA_IF are shown in Fig. 3. In this Figure, each row represents the input and output images of one dataset (i.e., source image1, source image2, fused images of DWT, UDWT, DWT-GA, UDWT-GA in sequence from left to right). It has been shown for all the four datasets.

The fused images obtained are validated with the objective metrics such as information entropy (IE), Universal Quality Index (QI), mutual information (MI), root mean square error (RMSE), spatial frequency (SF) and peak signal to noise ratio (PSNR), chosen as given in Sect. 3. The values observed are tabulated in Table 2. For readability, we have notated the methods as A1 for DWT_IF, A2 for UDWT_IF, A3 for DWT-GA_IF, A4 for UDWT-GA_IF.

From the subjective evaluation (i.e., human perception), we can justify that the contrast and tumour region are shown better in GA-based DWT fusion techniques than the DWT techniques without GA, where the contrast is retained in the outer edge levels of the fused image.

AscanbeseenintheTable2,theInformationEntropy(IE) values are considerably high for UDWT_IF (A2) algorithm, since only upsampling is carried out in decomposition stage.

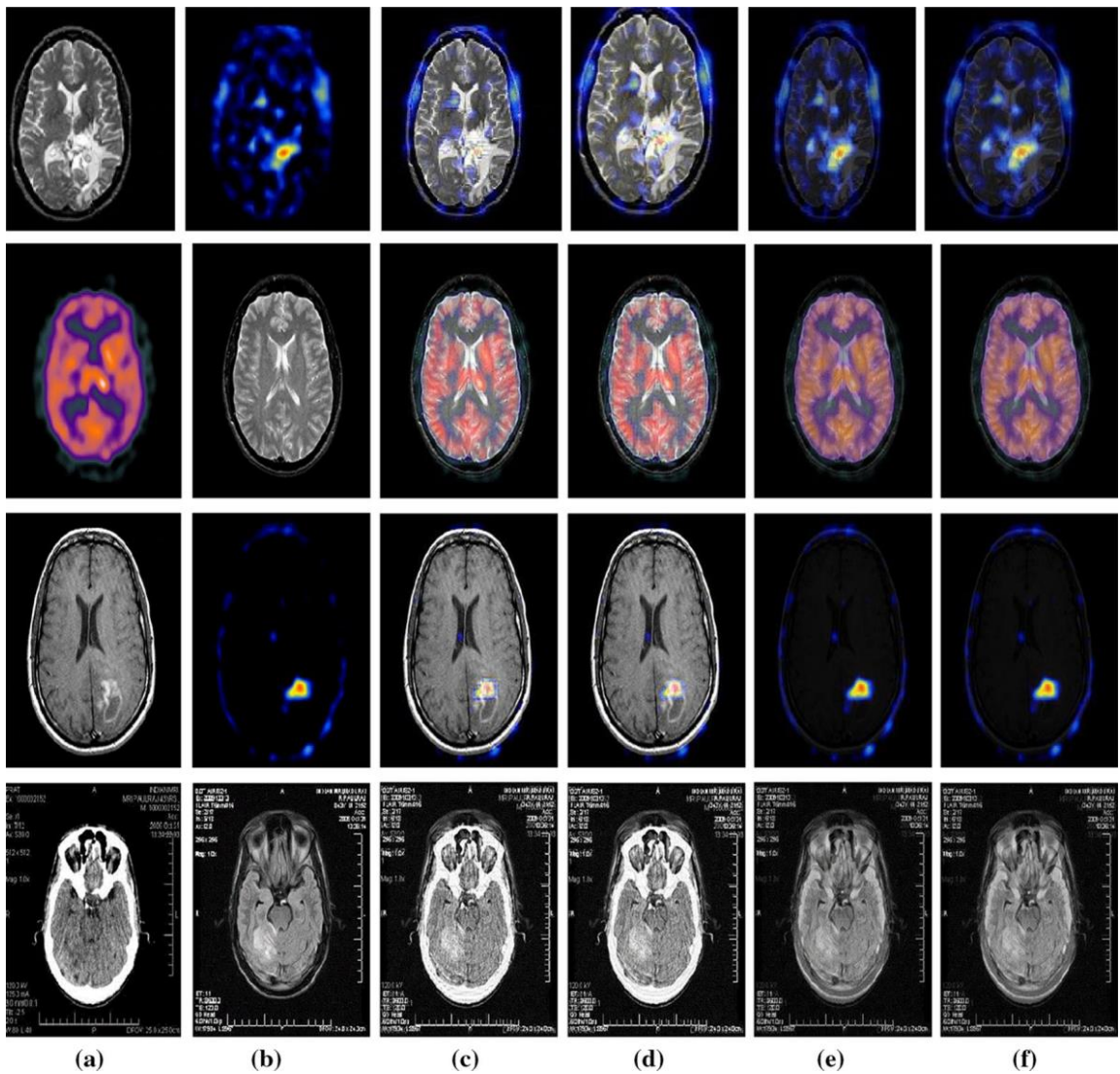

**Fig. 3** Input and output images of all the dataset Vs techniques. **a** Source image1, **b** Source image2, **c** DWT_IF, **d** UDWT_IF, **e** DWT-GA_IF, **f** UDWT-GA_IF (Dataset 1 to Dataset 4)

This in turn retains the shift invariance and gives increased amount of information about the transformed signal compared to the DWT. The number of the wavelet coefficients does not shrink between the transform levels. This additional information can be very useful for better analysis and understanding of the signal properties. Here signal is transformed into pixel intensities. The comparison of IE values across all datasets, shows that it is less only for dataset3 (Spect–Ti with MR-Gad), as one of the source images is MR-Gad which is generally a low-contrast image. The deviation of fused to source images (mutual information) is higher for DWT-GA and UDWT-GA compared to DWT and UDWT. RMSE is reduced and PSNR is increased across all the datasets, in optimal weight value-based GA only. QI metric depends on correlation, luminance and contrast. In A3, correlation factor is somewhat higher than A4 whereas luminance and contrast remain the same.

Table 2 Computed objective metric values for image set vs fusion technique

| Objective metric | Algorithm | Image set1 | Image set2 | Image set3 | Image set4 |
|---|---|---|---|---|---|
| IE | A1 | 4.4585 | 4.5443 | 5.0659 | 6.3205 |
|  | A2 | 4.4634 | 4.5588 | 5.0677 | 6.3409 |
|  | A3 | 4.0930 | 4.4399 | 3.5044 | 6.3388 |
|  | A4 | 4.0912 | 4.4402 | 3.4981 | 6.3378 |
| MI | A1 | 1.1743 | 1.4420 | 1.0979 | 1.2965 |
|  | A2 | 1.1668 | 1.4371 | 1.0950 | 1.2676 |
|  | A3 | 1.2251 | 1.4651 | 1.3047 | 1.3631 |
|  | A4 | 1.2260 | 1.4654 | 1.3049 | 1.3634 |
| RMSE | A1 | 6.8653 | 7.1786 | 6.6775 | 8.8554 |
|  | A2 | 6.8176 | 7.1262 | 6.6519 | 8.8106 |
|  | A3 | 6.2968 | 6.2752 | 4.7810 | 7.9097 |
|  | A4 | 6.2959 | 6.2755 | 4.7373 | 7.9050 |
| PSNR | A1 | 16.3200 | 19.4313 | 19.1628 | 13.2956 |
|  | A2 | 16.4429 | 19.6522 | 19.4408 | 13.4090 |
|  | A3 | 18.2951 | 21.9500 | 20.2696 | 16.5049 |
|  | A4 | 18.3122 | 21.9507 | 20.3187 | 16.4774 |
| QI | A1 | 0.6857 | 0.8638 | 0.6569 | 0.6307 |
|  | A2 | 0.6955 | 0.8616 | 0.6402 | 0.6769 |
|  | A3 | 0.6598 | 0.8731 | 0.5413 | 0.7201 |
|  | A4 | 0.6474 | 0.8542 | 0.5102 | 0.6801 |
| SF | A1 | 21.3960 | 20.1551 | 30.8098 | 79.1366 |
|  | A2 | 21.1865 | 19.5533 | 30.7513 | 78.7466 |
|  | A3 | 27.9144 | 31.9173 | 19.032 | 75.3817 |

Table 3: Optimal weight values (wv, wt) computed from genetic function for image set vs fusion techniques

| Algorithm | Optimal weight values from GA | Image set1 | Image set2 | Image set3 | Image set4 |
|---|---|---|---|---|---|
| A3 | wv | 0.2965 | 0.4772 | 0.0936 | 0.3201 |
|    | wt | 0.7035 | 0.5228 | 0.9064 | 0.6799 |
| A4 | wv | 0.2946 | 0.4763 | 0.0924 | 0.3238 |
|    | wt | 0.7054 | 0.5237 | 0.9076 | 0.6762 |

Therefore, QI is slightly higher in A3 than A4. When any one of the source images has high contrast (SPECT-Ti), the luminance factor dominates, thereby resulting in higher QI for basic wavelet fusion algorithms (A1, A2) than the proposed approach (A3, A4). Therefore, our proposed fusion process performs well for dataset2 and dataset4 which do not have SPECT-Ti as source images. From the SF metric value, it is inferred that the overall spatial quality depends on the contrast of the source images only, which can be observed through visual perception (refer Fig. 3). The contrast of the fused image for datasets 1 and 2 is less than that obtained for datasets 3 and 4 using the proposed approach. Therefore, the SF value is high for datasets 1 and 2 only. The same is reflected in weight computation also. also. When the difference in wv and wt is less, SF results with high value otherwise SF returns less value. Hence, the SF value of A3 is slightly higher than A4.

The computed weight values of DWT-GA and UDWT-GA techniques (A3, A4) are shown in Table 3. When comparing the weight values of source image1 (*wv*) and source image2 (*wt*), the source image with higher weight is observed to contribute more to the fused image. The weight values computed using both the techniques are in the same range. For dataset 3, the difference between *wv* and *wt* is higher since *wt* is derived from a low-contrast MRI- Gad image. From these weight values we also inferred that, while fusing the MRI and SPECT, the SPECT image contributes more to the fused image (w*t* > wv); and in the fusion of CT and MRI (dataset 4), the MRI image contributes more (w*t* > wv). The contrast, texture and tissue information are exactly transferred to the fused image from the source images which is also quantitatively validated.

**5 Conclusion and discussion**

The focus of this research work is to devise an algorithm for multimodality medical (brain) image fusion and to improve the quality of the fused image in terms of information, contrast and edge, and without false information or information loss. In addition to that, it should be possible to identify the source image which contributes more information to the fused image, preferably using lesser search space. To achieve these objectives, wavelet is combined with genetic algorithm (GA) for optimal parameter estimation and fusion process. The source images are decomposed using wavelet (DWT/UDWT) and from each region the features are extracted and applied to a genetic function. The number of features remains the same irrespective of the size of the image. From the genetic function, the optimal weight value of each source image is identified and the fusion process is carried out. In the genetic function, MSE is used as fitness criteria and the randomness in selection of initial population, is eliminated using features extracted after decomposition. Also, the obtained fused images using DWT, UDWT, DWT-GA, UDWT-GA fusion techniques are analyzed using subjective and objective metrics (IE, QI, MI, RMSE, PSNR, SF). Here subjective evaluation (qualitative) is a human perception whereas objective evaluation (quantitative) is a statistical computation.

From the results of quantitative evaluation, it is observed that, the MI value increases significantly with lesser RMSE and high PSNR for algorithms A3 and A4 (wavelet with GA) across all datasets. This indicates that complementary information is merged without information loss or false information. The IE values are considerably high for UDWT_IF (A2) algorithm, since only upsampling is carried out during the decomposition stage. QI metric depends on correlation, luminance and contrast. When any one of the source image has high contrast (SPECT-Ti), the luminance factor dominates, thereby resulting in higher QI for basic wavelet fusion algorithms (A1, A2) than the proposed approach (A3, A4). Therefore, our proposed fusion process performs well for dataset2 and dataset4 which do not have SPECT-Ti as source images. From the SF metric value it is inferred that the overall spatial quality depends on the contrast of the source images only, which can be observed through visual perception. The contrastofthefusedimagefordatasets1and2islessthanthat obtained for datasets 3 and 4 using the proposed approach. Therefore, the SF value is high for datasets1 and 2 only. The same is reflected in weight computation also. When the difference in $wv$ and $wt$ is less, it results in a high value of SF, else SF remains low. For dataset 3, the difference between $wv$ and $wt$ is high since $wt$ is derived from a low-contrast MRI-Gad image. From these weight values we also infer that, while fusing the MRI and SPECT, SPECT image contributes more to the fused image ($wt > wv$). In the fusion of CT and MRI (dataset 4), MRI image contributes more to the fused image ($wt > wv$). Thus the contrast, texture and tissue information are exactly transferred to the fused image from the source images is also quantitatively validated. The same algorithms can be used to fuse '$n$' source images also. Even though improvements in the fused image are obtained from the optimal weight estimation with complexity lessthan (Majidetal.2008),classification-based applications can be developed to ensure the importance of fused image set than the single modality for computer-aided disease diagnosis.